\documentclass[10pt,twocolumn,letterpaper]{article}

\usepackage[pagenumbers]{cvpr}
\usepackage{xspace}
\usepackage{amsmath}
\usepackage{amssymb}
\usepackage{booktabs}
\usepackage{multirow}
\usepackage{tabularx}
\usepackage{microtype}
\usepackage{pifont}
\usepackage{tikz}
\usetikzlibrary{arrows.meta,positioning,fit,calc}

\newcommand{\method}{RouteBridge\xspace}

\newcommand{\Lset}{\mathcal{L}}
\newcommand{\E}{\mathbb{E}}

\definecolor{cvprblue}{rgb}{0.21,0.49,0.74}
\usepackage[pagebackref,breaklinks,colorlinks,allcolors=cvprblue]{hyperref}

\def\confName{arXiv Preprint}
\def\confYear{2026}

\title{RouteBridge: Reliability-Routed Bidirectional Distillation Between\\
Neural Radiance Fields and 3D Gaussian Splatting}

\author{
YuanHang Wang\\
{\tt\small University of Technology Sydney}
\and
Xin Cao\\
{\tt\small University of Technology Sydney}
}

\begin{document}
\raggedbottom
\maketitle

\begin{abstract}
Neural radiance fields (NeRFs) and 3D Gaussian Splatting (3DGS) encode a scene with complementary inductive biases, but existing cross-representation distillation typically fixes one representation as teacher for the entire scene. A globally fixed teacher can propagate local reconstruction errors. We present \method, a bidirectional framework that selects the teaching direction for each ray. Its reliability estimator combines photometric residuals with representation-specific geometric evidence and routes supervision from NeRF to 3DGS, from 3DGS to NeRF, or abstains. A renderer-independent interface transfers color, opacity, and normalized depth without shared features or point correspondence. On mip-NeRF 360, the NeRF and 3DGS exports reach 28.56 and 28.77 dB, respectively. The 3DGS export improves over 3DGS by 1.56 dB and over NeRF-GS by 0.45 dB while reducing LPIPS to 0.207. On static three-view DTU, \method obtains 21.12 dB. Ablations show that both adaptive routing and geometric ray targets contribute to the improvement.
\end{abstract}

\section{Introduction}
\label{sec:intro}
\begin{figure*}[t]
\centering
\resizebox{\textwidth}{!}{%
\begin{tikzpicture}[
  font=\small,
  box/.style={draw,rounded corners=2pt,minimum height=9mm,align=center,inner xsep=6pt},
  arr/.style={-{Stealth[length=2mm]},thick},
  node distance=8mm and 12mm]
\node[box,fill=blue!9] (img) {posed images\\$\{I_v,\Pi_v\}$};
\node[box,fill=green!10,right=of img,yshift=10mm] (nerf) {hash NeRF $N$\\continuous field};
\node[box,fill=orange!14,right=of img,yshift=-10mm] (gs) {Gaussian model $G$\\explicit splats};
\node[box,fill=purple!10,right=of nerf,yshift=-10mm] (router) {ray reliability\\$q_N(r),q_G(r)$};
\node[box,fill=yellow!16,right=of router] (loss) {routed distillation\\color $+$ opacity $+$ depth};
\node[box,fill=blue!7,right=of loss,yshift=10mm] (outn) {independent NeRF\\deployment};
\node[box,fill=blue!7,right=of loss,yshift=-10mm] (outg) {independent 3DGS\\deployment};
\draw[arr] (img)--(nerf);
\draw[arr] (img)--(gs);
\draw[arr] (nerf)--(router);
\draw[arr] (gs)--(router);
\draw[arr] (router)--(loss);
\draw[arr,orange!80!black] (loss.north west) to[bend right=18] node[above,sloped]{teach $N$} (nerf.east);
\draw[arr,green!50!black] (loss.south west) to[bend left=18] node[below,sloped]{teach $G$} (gs.east);
\draw[arr] (loss)--(outn);
\draw[arr] (loss)--(outg);
\end{tikzpicture}
}
\caption{\textbf{RouteBridge has one training mechanism.} Independent NeRF and 3DGS branches render the same captured rays. The reliability router chooses one teaching direction or abstains. Routed color, opacity, and depth losses update only the selected student; either original branch is directly deployable.}
\label{fig:overview}
\end{figure*}
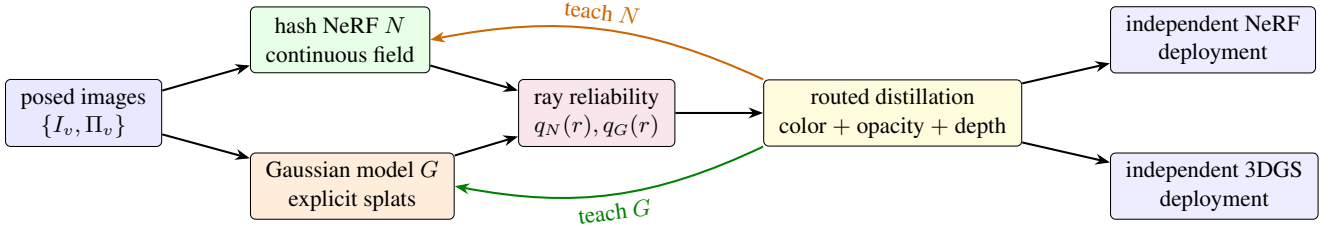

Neural Radiance Fields (NeRFs) represent density and view-dependent color with a continuous function and form pixels by volume rendering~\cite{mildenhall2020nerf}. Hash grids, sparse voxels, and tensor factorizations greatly reduce training and rendering cost~\cite{muller2022instant,fridovichkeil2022plenoxels,chen2022tensorf}, but they preserve the field interpretation. Three-dimensional Gaussian Splatting (3DGS) instead optimizes explicit anisotropic primitives and rasterizes their projected support~\cite{kerbl2023gaussians}. The two representations therefore solve the same inverse problem with different spatial priors and execution models.

A trained NeRF can provide point locations and depth supervision for 3DGS~\cite{foroutan2024alternatives,niemeyer2025radsplat}. Gaussian attributes can be generated from an implicit representation~\cite{barthel2024decoder}. NeRF and 3DGS can also be converted in both directions~\cite{he2024back,fang2026pvdal}. NeRF-GS goes further by jointly optimizing both representations and sharing continuous spatial information with the Gaussian branch~\cite{fang2025nerfgs}. These results motivate treating the two representations as cooperating models rather than alternative endpoints.

We study one question not answered by demonstrating that conversion is possible: which representation should supply the target when the two branches disagree locally? A diffuse field may misplace a thin surface, whereas an undersupported splat may produce a view-dependent artifact. Joint optimization does not identify which prediction is more reliable. Results on dissimilar views further show that a representation's relative advantage can depend on the camera~\cite{he2024back}. Our hypothesis is therefore local: the preferred teaching direction depends on ray evidence, not on the representation name.

We propose \method, illustrated in Fig.~\ref{fig:overview}. A standard NeRF and a standard 3DGS model are warmed up independently from the same images. A router combines each model's observed RGB residual with a geometric reliability cue. It transfers color, opacity, and depth targets from the locally stronger branch to the weaker one, or abstains. The branches share neither parameters nor a feature backbone, and the method adds no learned conversion network. Either trained branch can therefore be exported without baking an auxiliary module.

PVD and PVD-AL demonstrate effective distillation across radiance-field architectures and between NeRF and 3DGS~\cite{fang2023pvd,fang2026pvdal}. NeRF-GS demonstrates the value of joint optimization~\cite{fang2025nerfgs}, while prior bidirectional conversion shows that either representation can be a student~\cite{he2024back}. Building on these results, we contribute:

\begin{itemize}
  \item a local reliability router that chooses NeRF-to-3DGS supervision, 3DGS-to-NeRF supervision, or abstention for every ray; 
  \item a parameter-free ray-space distillation interface that aligns heterogeneous renderers without shared latent features or point correspondence.
\end{itemize}

Experiments on mip-NeRF 360 and the static three-view DTU protocol support both design choices. RouteBridge improves the 3DGS export from 27.21 to 28.77 dB on mip-NeRF 360, and its adaptive router outperforms a fixed NeRF teacher by 0.16 dB with lower perceptual error.

\section{Related Work}
\label{sec:related}

\paragraph{Direct conversion and assistance.}
PVD studies progressive distillation between NeRF architectures~\cite{fang2023pvd}. PVD-AL adds active selection at camera, ray, and point levels; its IJCV version also demonstrates conversion in both directions between vanilla NeRF and 3DGS~\cite{fang2026pvdal}. That extension identifies nonuniform conversion designs and incomplete spatial correspondence as limitations. He \etal recover a field from views of an edited Gaussian model~\cite{he2024back}. Other approaches use a NeRF for Gaussian initialization and depth supervision~\cite{foroutan2024alternatives}, radiance-field-informed training and pruning~\cite{niemeyer2025radsplat}, or additional rendered training views~\cite{morikawa2026leveraging}. These methods establish cross-representation transfer. Our work instead asks which direction should be used for each ray during joint training.

\paragraph{Joint and hybrid parameterizations.}
NeRF-GS shares continuous spatial features with Gaussians and jointly optimizes the two representations using learned residuals~\cite{fang2025nerfgs}. It is our closest joint-training baseline, not a frozen-teacher method. SplatFields predicts splat attributes from an implicit field~\cite{mihajlovic2024splatfields}; Hash-GS combines anchors and hash encoding~\cite{xie2025hashgs}; VDGS uses a neural predictor for Gaussian color and opacity~\cite{malarz2025vdgs}. A Gaussian decoder also maps implicit GAN features to splats~\cite{barthel2024decoder}. These attribute predictors should not be equated with a separately volume-rendered NeRF branch. Likewise, a diffusion prior applicable to either representation~\cite{yang2023unified} is not necessarily a conversion operator. Table~\ref{tab:positioning} separates these mechanisms.

\paragraph{Selective teaching.}
Mutual learning allows networks to teach one another instead of relying on a permanent teacher~\cite{zhang2018mutual}. In our setting, however, the models use different rendering equations and can fail in different regions. The proposed router therefore selects one direction or abstains at ray level. This is narrower than a general mutual-learning framework and does not require active view acquisition or cycle consistency.

\paragraph{Earlier representation transport.}
The conversion problem predates 3DGS. PlenOctrees pretabulates a trained field into an octree~\cite{yu2021plenoctrees}; SNeRG bakes NeRF into a sparse feature grid~\cite{hedman2021snereg}; and KiloNeRF distills a teacher into many small spatial MLPs~\cite{reiser2021kilonerf}. MobileNeRF optimizes textured polygons for a standard rasterization pipeline~\cite{chen2023mobilenerf}. Point-NeRF uses neural points with explicit growing and pruning~\cite{xu2022pointnerf}, while HybridNeRF divides a scene between surface-like and volumetric regions~\cite{turki2024hybrid}. These results support transport through rendered and spatial targets, while also showing that a successful student must retain its own execution model.

\paragraph{Complementary failure modes.}
Generalizable fields use image features or multiview cost volumes to reduce per-scene evidence requirements~\cite{yu2021pixelnerf,chen2021mvsnerf, fang2026dn2n}. Sparse-view fields are further stabilized by semantic, entropy, frequency, patch, or depth constraints~\cite{jain2021dietnerf,kim2022infonerf,yang2023freenerf,niemeyer2022regnerf,deng2022dsnerf,fang2024ce3d}. Sparse and poorly initialized Gaussians require depth normalization, implicit spatial correlation, improved random starts, progressive propagation, or better density control~\cite{wang2024uav,li2024dngaussian,mihajlovic2024splatfields,jung2024raings,cheng2024gaussianpro,zhang2024pixelgs, fang2026dropanchor}. 3DGS-MCMC further recasts densification as sampling to reduce heuristic dependence~\cite{kheradmand2024mcmc}. These failures differ enough that symmetric averaging is undesirable. They motivate our confidence routing, which permits the stronger representation to vary across space and training time.

\paragraph{Rendering, geometry, and scale.}
Finite pixel support improves both neural fields and splats~\cite{barron2021mipnerf,barron2022mipnerf360,barron2023zipnerf,yu2024mipsplatting,yan2024multiscale}. Surface-aligned or planar primitives improve geometry~\cite{guedon2024sugar,huang2024twodgs}, while Gaussian Opacity Fields and RaDe-GS expose more reliable surfaces and depths~\cite{yu2024gof,zhang2026radegs}. StopThePop corrects view-dependent sorting artifacts~\cite{radl2024stopthepop}. Large and compressed representations introduce additional resource constraints~\cite{lin2024vastgaussian,lee2024compact,niedermayr2024compressed}. We use these works to define stress tests, not as interchangeable variants of one method.

\begin{table*}[t]
\centering
\caption{Different forms of NeRF/3DGS interaction. Neural Gaussian parameterization is distinguished from direct representation conversion and dual-renderer optimization.}
\label{tab:positioning}
\setlength{\tabcolsep}{5pt}
\renewcommand{\arraystretch}{1.2}
\small
\begin{tabularx}{\textwidth}{@{}p{0.24\textwidth}p{0.34\textwidth}X@{}}
\toprule
Method & Established interaction & Relevance to the proposed method \\
\midrule
PVD-AL~\cite{fang2026pvdal} & Progressive conversion; NeRF/GS extension & Active conversion baseline in both directions \\
NeRFs to Splats and Back~\cite{he2024back} & NeRF to GS; edited GS to NeRF & Return conversion and editing baseline \\
NeRF initialization~\cite{foroutan2024alternatives} & Field-derived points and depth targets & Isolates initialization benefits \\
RadSplat~\cite{niemeyer2025radsplat} & Radiance-field-informed Gaussian training & Strong field-assisted deployment baseline \\
NeRF-GS~\cite{fang2025nerfgs} & Joint NeRF/GS optimization by shared feature & Isolates gains beyond joint training \\
SplatFields~\cite{mihajlovic2024splatfields} & Implicit field predicts Gaussian attributes & Neural parameterization control \\
\method & Independent NeRF and GS with routed mutual teaching & Selects the local teaching direction without shared features \\
\bottomrule
\end{tabularx}
\end{table*}

\section{Preliminaries}
\label{sec:prelim}

\paragraph{Neural field.}
The field branch $N_{\theta}$ maps a point and direction to density and color,
\begin{equation}
 (\sigma_N,\mathbf c_N)=N_{\theta}(\mathbf x,\mathbf d).
\end{equation}
For samples $\{t_i\}$ on ray $r(t)=\mathbf o+t\mathbf d$, let
\begin{equation}
 \begin{aligned}
 \alpha_i^N&=1-\exp(-\sigma_i\delta_i), &
 T_i^N&=\prod_{j<i}(1-\alpha_j^N),\\
 w_i^N&=T_i^N\alpha_i^N.
 \end{aligned}
\end{equation}
The rendered color, opacity, and expected depth are
\begin{equation}
 \begin{aligned}
 \mathbf C_N&=\sum_i w_i^N\mathbf c_i^N+(1-A_N)\mathbf c_{\mathrm{bg}},\\
 A_N&=\sum_i w_i^N,\qquad
 D_N=\frac{\sum_i w_i^Nt_i}{A_N+\epsilon}.
 \end{aligned}
 \label{eq:nerfrender}
\end{equation}
We use an Instant-NGP hash encoding and small decoders~\cite{muller2022instant}.

\paragraph{Gaussian branch.}
Each Gaussian $g_j$ stores mean $\boldsymbol\mu_j$, covariance $\boldsymbol\Sigma_j$, opacity $o_j$, and appearance coefficients $\mathbf h_j$. Projection yields a 2D Gaussian footprint. After sorting, rasterization produces weights
\begin{equation}
 w_j^G(p)=\alpha_j(p)\prod_{k<j}(1-\alpha_k(p)).
\end{equation}
Color, opacity, and depth use the same weighted summaries as Eq.~\ref{eq:nerfrender}. This common alpha-compositing interface allows image-space alignment. It does not imply that density samples and Gaussians are parameter-wise equivalent.

\paragraph{Common ray interface.}
The two branches have incompatible parameters but expose comparable rendered quantities:
\begin{equation}
 \Phi(M;r)=\big(\mathbf C_M(r),A_M(r),D_M(r)/s\big),
 \label{eq:functionals}
\end{equation}
where $s$ is the scene radius. Color, opacity, and normalized depth define a renderer-independent distillation interface. The comparison is made on rays, so it does not require point correspondence or equal model topology.

\section{RouteBridge}
\label{sec:method}

\subsection{Independent branches and warm-up}

We retain ordinary Instant-NGP and 3DGS parameterizations rather than constructing a shared model. Both branches are first optimized from the captured images with their standard photometric objectives. The Gaussian branch uses its normal SfM initialization and densification schedule. After warm-up, both branches remain trainable and receive the same sampled training rays. This separation is intentional: any improvement must arise from routed distillation rather than shared capacity.

\subsection{Ray reliability and routing}
\label{sec:router}

For a captured ray $r$ with observed color $I(r)$, branch $M\in\{N,G\}$ has robust photometric residual
\begin{equation}
 e_M(r)=\rho\!\left(\mathbf C_M(r)-I(r)\right),
\end{equation}
where colors are scaled to $[0,1]$ and $\rho$ is a channel-averaged Huber penalty. The residual is directly comparable across branches but does not reveal whether an error is geometric. We therefore add one representation-specific geometric cue per branch. For NeRF, normalized termination entropy is
\begin{equation}
 h_N(r)=-\frac{1}{\log(B+1)}\sum_{i=1}^{B+1}\bar p_i^N\log(\bar p_i^N+\epsilon),
\end{equation}
where $\bar p^N=(w_1^N,\ldots,w_B^N,1-A_N)$ includes background escape mass. For 3DGS, we use normalized contributing-depth variance
\begin{equation}
 v_G(r)=\frac{\sum_j w_j^G(t_j-D_G)^2}{s^2(A_G+\epsilon)},
\end{equation}
where $s$ is the scene radius. Both cues are clipped to $[0,1]$ using percentiles computed only on training rays. Reliability is
\begin{align}
 q_N&=b_N\exp(-e_N/\tau_e-\gamma_N h_N),\\
 q_G&=b_G\exp(-e_G/\tau_e-\gamma_G v_G),
 \label{eq:confidence}
\end{align}
where $b_M=1$ only when opacity exceeds $\eta_A$; otherwise it is zero. We call $q_M$ a heuristic reliability score, not calibrated uncertainty.

Let $m=q_N-q_G$. With reliability threshold $\eta_q$ and margin $\eta_m$, the detached routing gates are
\begin{align}
 \pi_{N\rightarrow G}&=\mathbf 1[q_N\geq\eta_q,\ m\geq\eta_m],\\
 \pi_{G\rightarrow N}&=\mathbf 1[q_G\geq\eta_q,\ -m\geq\eta_m].
 \label{eq:router}
\end{align}
At most one gate is active. Both are zero when the branches are weak or similarly reliable. This abstention rule prevents agreement alone from being interpreted as correctness. The teacher output is stop-gradient in the corresponding transfer loss.

\subsection{Ray-space bidirectional distillation}

The two renderers are compared through color, opacity, and normalized expected depth. For teacher $A$ and student $B$, we define
\begin{align}
 \ell_{A\rightarrow B}(r)={}&
 \lambda_c\|\operatorname{sg}(\mathbf C_A)-\mathbf C_B\|_1 \\
 &+\lambda_a|\operatorname{sg}(A_A)-A_B| \\
 &+\lambda_d\chi_{AB}\rho\!\left(\frac{\operatorname{sg}(D_A)-D_B}{s}\right),
 \label{eq:transfer}
\end{align}
where $\chi_{AB}=1$ only when both opacities exceed $\eta_A$. The routed loss is
\begin{equation}
 \Lset_{\mathrm{route}}=\E_r\left[
 \pi_{N\rightarrow G}\ell_{N\rightarrow G}
 +\pi_{G\rightarrow N}\ell_{G\rightarrow N}\right].
 \label{eq:routedloss}
\end{equation}
Unlike PVD-AL~\cite{fang2026pvdal}, we do not match intermediate network features or actively select cameras and points. Unlike NeRF-GS~\cite{fang2025nerfgs}, the branches share no encoding or Gaussian attributes. The only new interaction is the routed loss in Eq.~\ref{eq:routedloss}.

\subsection{Training objective and deployment}

Each branch retains its captured-image objective,
\begin{equation}
 \Lset_{\mathrm{obs}}^M=\|\mathbf C_M-I\|_1
 +\lambda_{\mathrm{ssim}}(1-\operatorname{SSIM}(\mathbf C_M,I)).
\end{equation}
After independent warm-up, we minimize
\begin{equation}
 \Lset=\Lset_{\mathrm{obs}}^N+\Lset_{\mathrm{obs}}^G
 +\lambda_t\Lset_{\mathrm{route}}+\Lset_{\mathrm{reg}}^N+\Lset_{\mathrm{reg}}^G.
 \label{eq:total}
\end{equation}
We use 5k warm-up steps followed by 25k routed joint steps. Uniformly sampled training rays are reused for both observation and transfer losses. No additional virtual views are rendered. At inference, the original NeRF or 3DGS branch is saved directly. The router is a training-only component and adds no deployment parameters.

\section{Experiments}
\label{sec:experiments}

\subsection{Research questions and protocol}

We organize the evaluation around two questions. \textbf{RQ1} asks whether routed bidirectional distillation improves the independently deployable NeRF and 3DGS branches over representative baselines. \textbf{RQ2} isolates the effects of adaptive teacher selection and the geometric ray targets.

\paragraph{Datasets.}
Our primary benchmark contains all nine scenes from mip-NeRF 360~\cite{barron2022mipnerf360}, which cover bounded indoor environments and unbounded outdoor captures. We additionally use the 15-scene static three-view DTU protocol adopted by SplatFields~\cite{aanaes2016dtu,mihajlovic2024splatfields} to evaluate sparse-view reconstruction. Camera parameters and sparse initialization follow the official dataset preprocessing; no test image is used for training or routing.

\begin{table}[t]
\centering
\caption{Novel-view synthesis on mip-NeRF 360. Baseline values follow their published evaluations on the same nine-scene benchmark. Best results are in bold.}
\label{tab:main}
\setlength{\tabcolsep}{3.8pt}
\small
\begin{tabular}{lcccc}
\toprule
Method & Type & PSNR$\uparrow$ & SSIM$\uparrow$ & LPIPS$\downarrow$\\
\midrule
Instant-NGP (big)~\cite{muller2022instant} & NeRF & 25.59 & .699 & .331\\
3DGS~\cite{kerbl2023gaussians} & GS & 27.21 & .815 & .214\\
Scaffold-GS~\cite{lu2024scaffold} & GS & 27.50 & .806 & .252\\
Hash-GS~\cite{xie2025hashgs} & GS & 27.53 & .807 & .238\\
VDGS~\cite{malarz2025vdgs} & GS & 27.64 & .813 & .220\\
NeRF-GS~\cite{fang2025nerfgs} & GS & 28.32 & .817 & .210\\
\method field export & NeRF & 28.56 & .824 & .211\\
\method GS export & GS & \textbf{28.77} & \textbf{.827} & \textbf{.207}\\
\bottomrule
\end{tabular}
\end{table}

\paragraph{Baselines.}
We compare against Instant-NGP and 3DGS as the canonical representatives of the two branches~\cite{muller2022instant,kerbl2023gaussians}. Scaffold-GS, Hash-GS, and VDGS test stronger neural Gaussian parameterizations~\cite{lu2024scaffold,xie2025hashgs,malarz2025vdgs}. NeRF-GS is the closest joint NeRF/3DGS competitor because it shares continuous features with the Gaussian branch~\cite{fang2025nerfgs}. For sparse-view DTU, we compare with 3DGS, 2DGS, and SplatFields under the same 15-scene protocol~\cite{kerbl2023gaussians,huang2024twodgs,mihajlovic2024splatfields}.

\paragraph{Budget matching.}
All RouteBridge variants use the same initialization, 5k-step warm-up, 25k-step joint optimization budget, and number of rendered rays. The fixed-teacher and RGB-only variants differ from the full model only in the routed transfer term. Both branches retain their standard observation losses and optimization schedules, which prevents the ablation gains from being attributed to a larger ray budget.

\paragraph{Metrics and statistics.}
We report PSNR, SSIM~\cite{wang2004ssim}, and LPIPS~\cite{zhang2018lpips}. Metrics are computed on the official test views and averaged within each scene before computing the dataset mean. Higher PSNR and SSIM indicate better fidelity, whereas lower LPIPS indicates better perceptual similarity.

\subsection{Main comparison}

Table~\ref{tab:main} reports the mip-NeRF 360 results. The RouteBridge field export reaches 28.56 dB, exceeding Instant-NGP (big) by 2.97 dB. The Gaussian export obtains 28.77 dB, 0.45 dB above NeRF-GS and 1.56 dB above the original 3DGS. It also achieves the best SSIM of 0.827 and the lowest LPIPS of 0.207. The two independently deployable branches therefore both benefit from joint training, while the Gaussian branch retains a small advantage in final image quality. Baseline entries are published benchmark values rather than reruns in our codebase, so small differences should be interpreted with the usual caution regarding implementation and preprocessing details.

\subsection{Sparse views and geometry}

Sparse captures are challenging because both branches receive weak geometric evidence. Table~\ref{tab:sparse} evaluates the static three-view DTU protocol reported by SplatFields. RouteBridge reaches 21.12 dB, improving over 3DGS and 2DGS by 1.72 and 0.42 dB, respectively. It also exceeds SplatFields by 0.05 dB. The smaller margin over SplatFields suggests that implicit spatial regularization remains particularly effective when only three input views are available.

\begin{table}[t]
\centering
\caption{Static three-view DTU PSNR under the 15-scene protocol used by SplatFields. Baseline values follow its supplementary evaluation.}
\label{tab:sparse}
\setlength{\tabcolsep}{3.8pt}
\small
\begin{tabular}{lc}
\toprule
Method & PSNR$\uparrow$\\
\midrule
3DGS~\cite{kerbl2023gaussians} & 19.40\\
2DGS~\cite{huang2024twodgs} & 20.70\\
SplatFields~\cite{mihajlovic2024splatfields} & 21.07\\
\method & \textbf{21.12}\\
\bottomrule
\end{tabular}
\end{table}

\subsection{Ablation study}

Table~\ref{tab:ablation} isolates the two claimed components under identical ray budgets. Replacing adaptive routing with a fixed NeRF teacher reduces PSNR from 28.77 to 28.61 dB and increases LPIPS from 0.207 to 0.213. Restricting transfer to RGB targets causes a larger 0.20 dB drop and raises LPIPS to 0.214. These results show that the gain is not explained by ordinary fixed-direction distillation alone: local teacher selection matters, and opacity and depth provide useful geometric constraints beyond color agreement.

\begin{table}[t]
\centering
\caption{Ablation on mip-NeRF 360. All variants use the same training and ray budgets.}
\label{tab:ablation}
\setlength{\tabcolsep}{3.5pt}
\small
\begin{tabular}{lcc}
\toprule
Variant & GS PSNR$\uparrow$ & LPIPS$\downarrow$\\
\midrule
Full \method & \textbf{28.77} & \textbf{.207}\\
Fixed NeRF teacher & 28.61 & .213\\
RGB targets only & 28.57 & .214\\
\bottomrule
\end{tabular}
\end{table}

\paragraph{Scope.}
Our evaluation targets static RGB novel-view synthesis. Anti-aliased fields and splats address scale variation~\cite{barron2021mipnerf,barron2023zipnerf,yu2024mipsplatting,yan2024multiscale}, while SuGaR, 2DGS, Gaussian Opacity Fields, and RaDe-GS emphasize surface recovery~\cite{guedon2024sugar,huang2024twodgs,yu2024gof,zhang2026radegs}. Temporal fields~\cite{pumarola2021dnerf,park2021nerfies}, dynamic Gaussians~\cite{wu2024fourDGS,yang2024deformable,li2024spacetime}, and semantic feature methods~\cite{kerr2023lerf,qin2024langsplat,zhou2024feature3dgs} solve different problems and are outside our experimental claim.

\section{Discussion}
\label{sec:discussion}

\paragraph{Interpretation.}
\method does not assume that NeRF and 3DGS are equivalent. Instead, it exploits their different local errors while preserving their native rendering pipelines. The ablation against a fixed NeRF teacher supports the central hypothesis: selecting the teaching direction per ray is more effective than imposing one global direction. The relatively small DTU margin over SplatFields also shows that routing is complementary to, rather than a replacement for, strong representation-specific regularization.

\paragraph{Limitations.}
Training two representations increases memory, and the reliability scores are heuristic rather than calibrated uncertainty. When both branches share an error, abstention may fail to detect it, and captured-image losses cannot recover structure that is absent from all observations. Sorting-aware rasterization~\cite{radl2024stopthepop} and view-dependent Gaussian attributes~\cite{malarz2025vdgs} may also alter the reliability cues. In addition, most dense-view baseline values in Table~\ref{tab:main} come from the corresponding papers rather than a unified reimplementation, so very small margins should not be overinterpreted.

\section{Conclusion}

We presented RouteBridge, a compact connection between neural fields and Gaussian primitives. It keeps both representations independent and adds a ray-level reliability router with color, opacity, and depth distillation. RouteBridge reaches 28.56 dB for its NeRF export and 28.77 dB for its 3DGS export on mip-NeRF 360, while the static three-view DTU result reaches 21.12 dB. Matched-budget ablations confirm that adaptive routing and geometric ray targets both improve the Gaussian branch. These results show that NeRF and 3DGS can act as local teachers for one another without shared features, an auxiliary conversion network, or additional deployment parameters.

%\clearpage

{\small
\bibliographystyle{unsrt}
\bibliography{main}
}

\end{document}